# TFT-ACB-XML: Decision-Level Integration of Customized Temporal Fusion Transformer and Attention-BiLSTM with XGBoost Meta-Learner for BTC Price Forecasting

Raiz Ud Din[1], Saddam Hussain Khan [2]*

[1]Artificial Intelligence Lab, Department of Computer Systems Engineering, University of Engineering and Applied Sciences (UEAS), Swat, Pakistan

[2]Interdisciplinary Research Center for Smart Mobility and Logistics, King Fahad University of Petroleum and Minerals (KFUPM), Dhahran, Saudi Arabia

**Email:** hengrshkhan822@gmail.com, saddam.khan@kfupm.edu.sa

## Abstract

Accurate forecasting of Bitcoin (BTC) has always been a challenge because decentralized markets are non-linear, highly volatile, and have temporal irregularities. Existing deep learning models often struggle with interpretability and generalization across diverse market conditions. This research presents a hybrid stacked-generalization framework, TFT-ACB-XML, for BTC closing price prediction. The framework integrates two parallel base learners: a customized Temporal Fusion Transformer (TFT) and an Attention-Customized Bidirectional Long Short-Term Memory network (ACB), followed by an XGBoost regressor as the meta-learner. The customized TFT model handles long-range dependencies and global temporal dynamics via variable selection networks and interpretable single-head attention. The ACB module uses a new attention mechanism alongside the customized BiLSTM to capture short-term sequential dependencies. Predictions from both customized TFT and ACB are weighted through an error-reciprocal weighting strategy. These weights are derived from validation performance, where a model showing lower prediction error receives a higher weight. Finally, the framework concatenates these weighted outputs into a feature vector and feeds the vector to an XGBoost regressor, which captures non-linear residuals and produces the final BTC closing price prediction. Empirical validation using BTC data from October 1, 2014, to January 5, 2026, shows improved performance of the proposed framework compared to recent Deep Learning and Transformer baseline models. The results show an MAPE of 0.65%, an MAE of 198.15, and an RMSE of 258.30 for one-step-ahead out-of-sample under a walk-forward evaluation on the test block. The evaluation period spans the 2024 BTC halving and the spot ETFs (Exchange-Traded funds) period, which coincide with major liquidity and volatility shifts.

**Keywords:** BTC; deep learning; TFT; BiLSTM; XGBoost; attention; transformer; price prediction.

# 1. Introduction

Cryptocurrencies have become an integral part of the financial industry and BTC (BTC) has disrupted traditional monetary systems since its launch in 2009 [1]. BTC is now recognized as a distinct asset class and had a market capitalization of $1.87 trillion out of a total crypto market of $3.13 trillion in mid-January 2026 [2]. The BTC's growing demand has increased its sensitivity to fluctuations due to global macroeconomic indicators, investor sentiment and liquidity dynamics [3]. Moreover, the BTC market then shifted regimes after the miner's block reward halving and the institutionalization of BTC through Spot Exchange-Traded Funds (ETFs) in January 2024. The halving created a supply shortage due to reducing mining rewards from 6.25 to 3.125 BTC [4]. Concurrently, institutional inflows through ETFs increased participation by large investors and raised BTC demand [5]. These changes altered the statistical behavior of the BTC price series and consequently making the BTC forecasting difficult. Therefore, more robust forecasting frameworks are required to handle these supply shocks and the evolving dynamics of institutional trading volume.

Traditionally, investors often rely on technical analysis to derive trading signals from historical price paths, using tools such as moving averages. Moreover, fundamental analysis evaluates underlying value through variables such as supply-demand imbalances and macroeconomic indicators [6]. However, both technical and fundamental analysis often struggle with the volatility in cryptocurrency markets because they are highly non-linear and speculative environments [7]. Consequently, researchers have turned to classical statistical methods for time series models to capture temporal dependencies [8].

Linear models such as ARIMA (autoregressive integrated moving average) models are effective in short- to medium-term forecasting tasks involving structured temporal data. However, these linear models work best with linear data but they struggle in capturing the stochastic and non-linear behaviors of cryptocurrency markets [9]. These challenges motivated the exploration of machine learning (ML) and deep learning (DL) strategies capable of capturing hidden temporal dependencies and non-linear relationships in cryptocurrency data.

A specialized DL architecture, LSTM (Long Short-Term Memory) has been effective in modeling price movements in cryptocurrency markets [10]. LSTM's gating mechanisms mitigate gradient decay and support sequential pattern extraction [11]. However, these models face difficulties in learning long-range dependencies and face scalability issues when modeling complex interactions among multiple market variables such as open, high, low, close, and volume (OHLCV). Therefore, transformer models widely used in natural language processing (NLP), large language models (LLMs) and medical diagnosis have also been introduced for sequential data to capture long-term dependencies [12], [13], [14], [15], [16], [17], [18], [19], [20], [21]. Moreover, combining LSTM with modern Transformer algorithms improves the modeling of local and global temporal dependencies in cryptocurrency forecasting [22]. Temporal Fusion Transformer (TFT) has the capability to learn known and unknown time-varying patterns and implement a temporal attention concept. These developments highlight a steady transition from traditional recurrent structures toward Transformer-based systems capable of handling extensive temporal relationships.

In this regard, this work proposes a novel hybrid stacked-generalization framework, TFT-ACB-XML, to address the aforementioned challenges in next-day BTC closing price prediction. The proposed framework consists of two parallel base learners; (i) a Temporal Fusion Transformer (TFT) and (ii) an Attention-Customized BiLSTM (ACB). The outcomes of these models are fed into a XGBoost based meta-learner. Moreover, a new attention mechanism in the ACB has been used in parallel with a customized BiLSTM to prioritize trading volume and daily closing prices and it enhances responsiveness to liquidity-driven fluctuations. The integration of these components in the proposed framework addresses both global context and local volatility. Evaluation uses a chronological train/validation/test split and a walk-forward test procedure with fixed model parameters and fixed weights during the test window. This study uses daily OHLCV variables as inputs to keep the setup reproducible and comparable. This study focuses on forecasting accuracy and robustness. The TFT architecture includes variable selection weights and temporal attention that offer qualitative insight into feature relevance and influential time steps. The main contributions of this study include the following:

- A novel hybrid stacked-generalization framework, TFT-ACB-XML, is proposed to learn heterogeneous temporal dynamics in BTC data. The framework targets robustness under regime changes by integrating a TFT and an ACB as parallel base learners, followed by non-linear refinement via an XGBoost meta-learner.

- The customized TFT model learns global temporal dependencies and the evaluation period includes the 2024 halving.

- The ACB module uses a new attention mechanism in parallel with a customized BiLSTM to capture sequential dependencies. The new attention mechanism prioritizes trading volume and price action to address short-term fluctuations and liquidity dynamics during the spot ETFs period.

- Independent outputs from two parallel base learners (customized TFT and ACB) are individually weighted through an error-reciprocal weighting mechanism. This weighting ensures that the meta-learner receives prioritized features based on the recent historical precision of each base model.

- These weighted outputs are then concatenated into a feature vector and fed into the XGBoost meta-learner. XGBoost evaluates the specific error patterns of each base learner to model non-linear residuals and correct inherent biases. This process produces the final BTC price prediction and improves the generalization and robustness of the TFT-ACB-XML framework.

- Experimental evaluation on daily BTC data (October 1, 2014 – January 5, 2026) shows lower MAPE, MAE and RMSE than the 15 compared sequential and transformer-based models under the reported walk-forward test setup.

The remaining sections are organized as follows. The next section presents the proposed TFT-ACB-XML framework section that explains the TFT-ACB-XML framework. Experimental Setup section outlines the experimental setup. The fourth section is the Result section that shows comparative evaluation with existing methods, supported by quantitative results. The last Conclusion section closes the study and highlights avenues for future work.

## 2. Literature Review

Forecasting cryptocurrency prices is hard due to non-linearity, volatility and noisy signals in the data. Classical statistical models such as ARIMA are effective for stable data over time but they struggle to handle the volatile nature of cryptocurrencies [23]. These limitations motivated the use of machine learning (ML) techniques for capturing non-linear relationships. Non-linear regressors such as kernel-based SVM, Random Forest and XGBoost are effective for finding complex patterns in data with minimal distributional requirements. However, these models rely on feature engineering to represent temporal memory and also their performance depends on the chosen lag structure and inputs [24].

Deep Learning (DL) architectures address several limitations of ML approaches and improve sequential modeling capability. Standard RNNs can suffer from vanishing gradients but gated architectures such as LSTM and Gated Recurrent Units (GRU) mitigate this issue and improve learning of sequential dependencies in time-series forecasting [25]. Bidirectional LSTM (BiLSTM) improves performance by processing the historical look-back window in both forward and backward directions. Integration of attention mechanisms in BiLSTM enables the model to focus on relevant historical time steps and prior studies report improved forecasting performance with attention-augmented BiLSTM [26], [27]. However, pure recurrent models are sensitive to abrupt price fluctuations and struggle with unfamiliar market conditions [28].

Hybrid models address these limitations by merging sequential learning with signal decomposition. Studies integrating BiLSTM-GRU, LSTM-GRU, and RNN integrated with VMD (Variational Mode Decomposition) and SE (Sample Entropy) reported improved forecasting efficiency [29], [30], [31], [32]. However, these designs increased architectural complexity and potential for error propagation during sudden market shifts. CNN–BiLSTM (CNNs - Convolutional Neural Networks) hybrids combine the convolutional feature extraction with the sequential learning and show better performance on a wide variety of evaluation metrics [33]. However, these models can struggle with volatility-induced changes in local patterns. Another study optimized BiLSTM with a customized attention mechanism and ensembled outputs with XGBoost (Attention Customized BiLSTM XGBoost Decision Ensemble- ACB-XDE) and achieved low MAPE of 0.37%, MAE of 84.40 and RMSE of 106.14 [34]. This study achieved a lower MAPE but its evaluation was limited to the 2014-2023 period. The proposed work extends the analysis to the 2014–2026 period and this period include a post-2024 regime with large market shifts. Direct metric comparison across studies remains limited because time windows, splits and evaluation protocols differ. Evaluation protocol influences reported accuracy, and walk-forward testing provides a realistic out-of-sample assessment for volatile markets. The proposed framework TFT-ACB-XML targets long-range dependency handling under this extended window. Recently, Transformer-based architectures and their time-series adaptations emerged as a widely used alternative. The self-attention mechanism allows models to capture long-range dependencies and cross-temporal interactions efficiently. A 2025 study introduced a decomposition-aware Transformer named Helformer for cryptocurrency price forecasting that combines Holt-Winters decomposition with attention layers and it outperformed LSTM, BiLSTM and GRU baselines on BTC data [35]. The authors report strong accuracy but the model relies mainly on price-based inputs and uses limited stress testing under abrupt price fluctuations. Similarly, another recent work adopted a Time-Series Transformer (TST) on BTC data and reported improved Mean Squared Error (MSE) and Mean Absolute Scaled Error (MASE) compared with LSTM during volatile periods [36]. The study highlights performance gains but uses a relatively small dataset and lacks comparison with residual correction or ensemble strategies. However, application of Transformers to cryptocurrency forecasting has shown mixed results. In the Helformer study, the vanilla Transformer model without decomposition produced higher RMSE and MAPE than the decomposition-based variant and showed sensitivity to raw volatility [35]. A 2025 preprint introduced a hybrid attention-Transformer + GRU model for BTC and Ethereum price prediction and it outperformed simple RNN and feed-forward baselines on MSE, RMSE, MAE and MAPE [37]. Furthermore, many existing hybrid Transformer-based frameworks omit validation-driven error-reciprocal weighting and a residual refinement, which can reduce predictive accuracy. Incorporating these techniques can help correct biases and improve forecasting accuracy.

**Research Gap**

Despite the evolution of DL models, several critical gaps remain in the literature:

- The abrupt price fluctuations and non-linear dynamics present considerable challenges for cryptocurrency forecasting.
- Many models focus either on short-term sequential dynamics or long-range dependencies but few frameworks effectively integrate both through a coordinated parallel architecture.
- Price is often the primary variable and the prioritization of correlation between price and liquidity-driven features (like trading volume) in attention mechanisms is often overlooked and leads to poor responsiveness during market shocks.
- Many hybrid models are end-to-end and do not implement a post-hoc residual refinement stage (like XGBoost) to correct the specific non-linear errors produced by the deep learning base.
- Standard ensembles use fixed averaging and these methods ignore varying sub-model performance across different market conditions.

The proposed TFT-ACB-XML framework addresses these limitations by jointly modeling both short-term and long-term temporal dependencies. This architecture prioritizes volume-driven price dynamics and applies adaptive residual correction to improve forecasting accuracy.

| No. | Study (year) | Model / Focus | Key result | Shortcomings |
|---|---|---|---|---|
| [35] | Kehinde et al. (2025) | Helformer: Holt-Winters + | Low RMSE/MAPE, strong | Relies mainly on price data and limited stress tests for |

| | | Transformer | transferability. | abrupt regime breaks. |
|---|---|---|---|---|
| [36] | Zheng (2025) | Time-Series Transformer (TST) | Improved MASE vs LSTM on BTC. | Small dataset, lacks residual refinement or ensemble strategies. |
| [38] | Zhou et al. (2021) | Informer (ProbSparse Attention) | Strong long-sequence forecasting. | Not crypto-specific, requires adaptation for high-noise financial data. |
| [39] | Wu et al. (2021) | Autoformer (Decomposition) | SOTA on long-term benchmarks. | High complexity. |
| [40] | Nie et al. (2023) | PatchTST (Channel Independence) | Efficient handling of long look-backs. | Not evaluated on volatile daily BTC series with liquidity features. |
| [41] | Liu et al. (2023) | iTransformer (Inverted Variates) | Strong multivariate performance. | Sensitivity to implementation details, not tested on crypto-market shocks. |
| [42] | Zeng et al. (2023) | DLinear (Linear vs. Transformer) | Linear baselines can beat Transformers. | Highlights that Transformers require specialized design for time-series. |

## 3. Proposed TFT-ACB-XML Framework

This study presents a hybrid stacked-generalization framework for daily BTC closing price prediction. The proposed framework, TFT-ACB-XML, targets abrupt changes in the statistical behavior of BTC prices. The evaluation period spans the 2024 BTC halving and the post-ETF approval period (2024), which coincide with major liquidity and volatility shifts. An overview of the proposed study methodology is shown in Figure 2. The proposed framework is assessed on a BTC dataset and benchmarked against other predictive models. The framework comprises two parallel base learners, (i) TFT and (ii) ACB, followed by an inverse error-based weighting strategy, feature concatenation, and a final XGBoost regressor that acts as a meta-learner. The weights are derived from validation performance only to preserve the integrity of the results and avoid data leakage.

The customized TFT is trained on multi-year historical data (covering multiple market regimes including the post-2024 period), while its model input uses a fixed look-back window (L = 60 days). In this paper, long-term refers to cross-regime relationships learned from the multi-year training history rather than a multi-year input sequence. The ACB targets short-term sequential dependencies and local volatility within the same historical look-back window. A new attention block is integrated with the BiLSTM of ACB to prioritize influential variables such as daily closing price and trading volume. Price and volume co-move in many market regimes and volume-related signals often coincide with liquidity shifts during the post-ETF approval period (2024). Both base learners produce independent predictions. These predictions are combined using inverse-error weighting computed on the validation split, where a lower validation error yields a higher weight. The weighted outputs are then concatenated into a feature vector and provided to the final XGBoost model. XGBoost uses gradient-boosted decision trees to model non-linear residual structure and reduce systematic errors left by the base learners. The TFT-ACB-XML framework is outlined architecturally in Figure 1 and its structural details are provided in Figure 3.

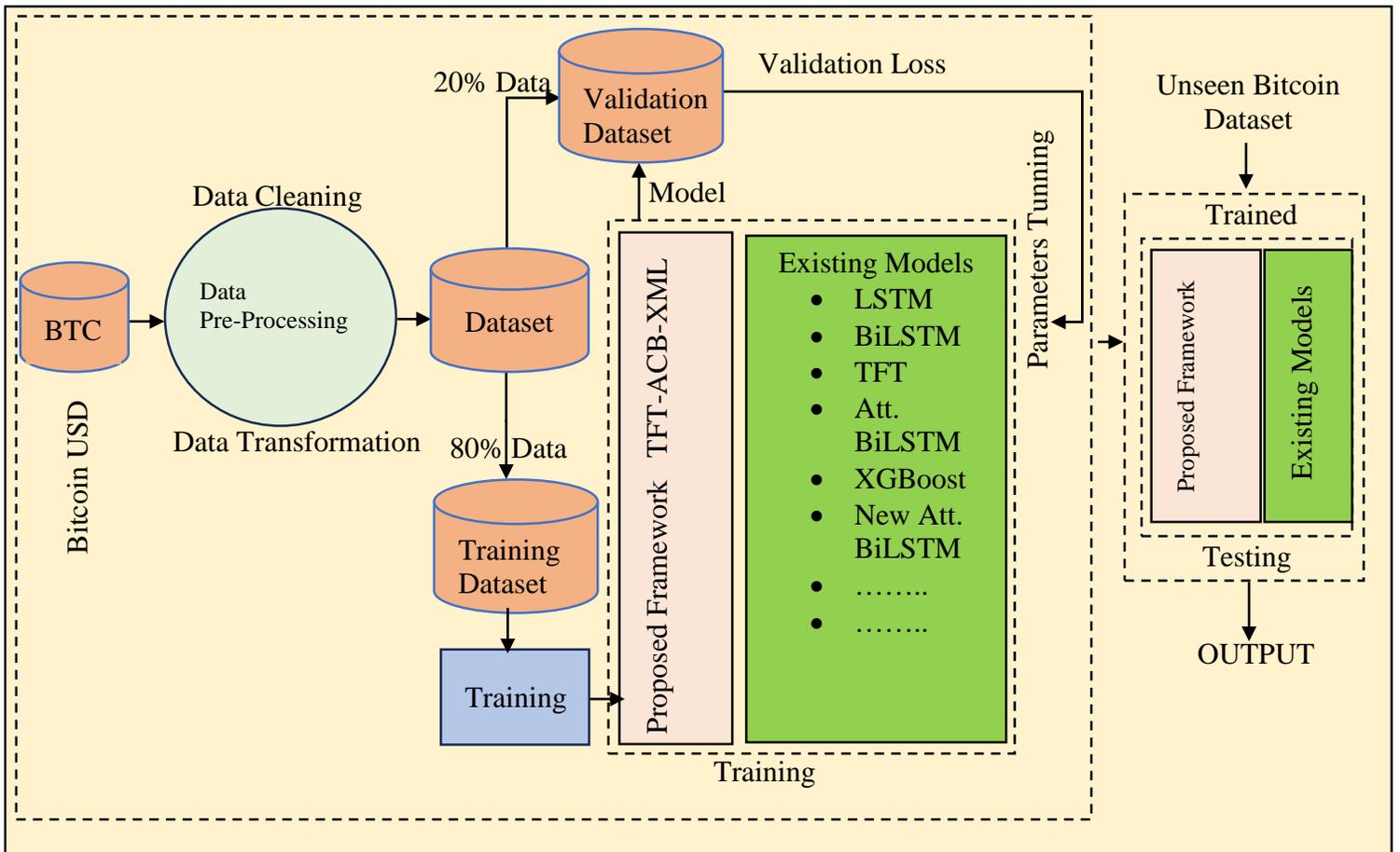

*Figure 2. Graphical overview of the proposed study.*

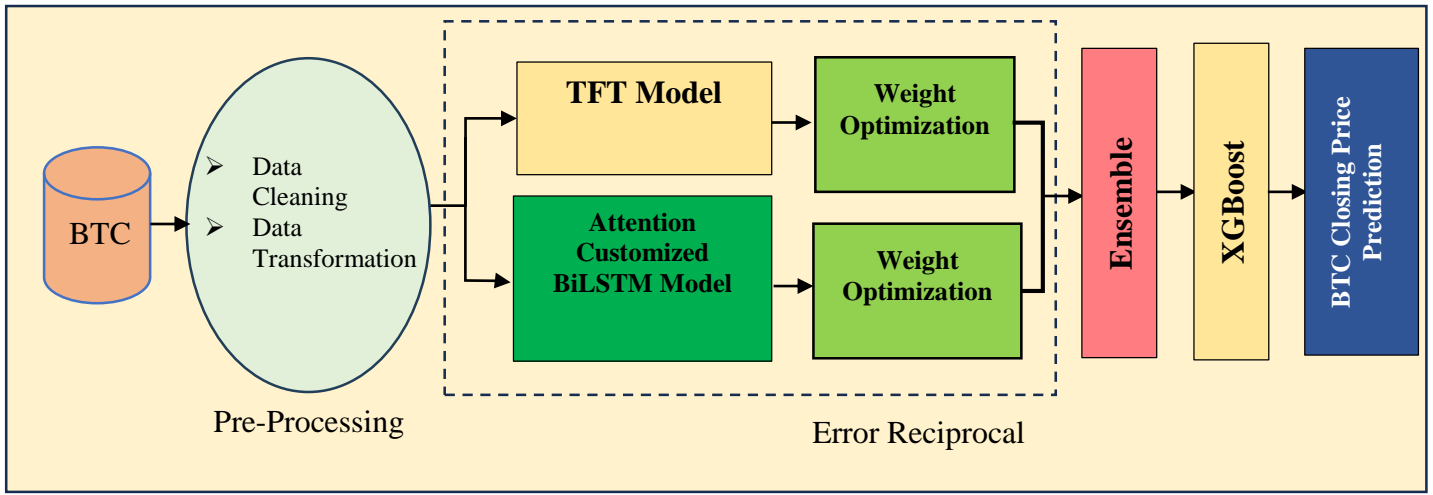

*Figure 1. Brief Overview of the proposed Bitcoin prediction TFT-ACB-XML framework.*

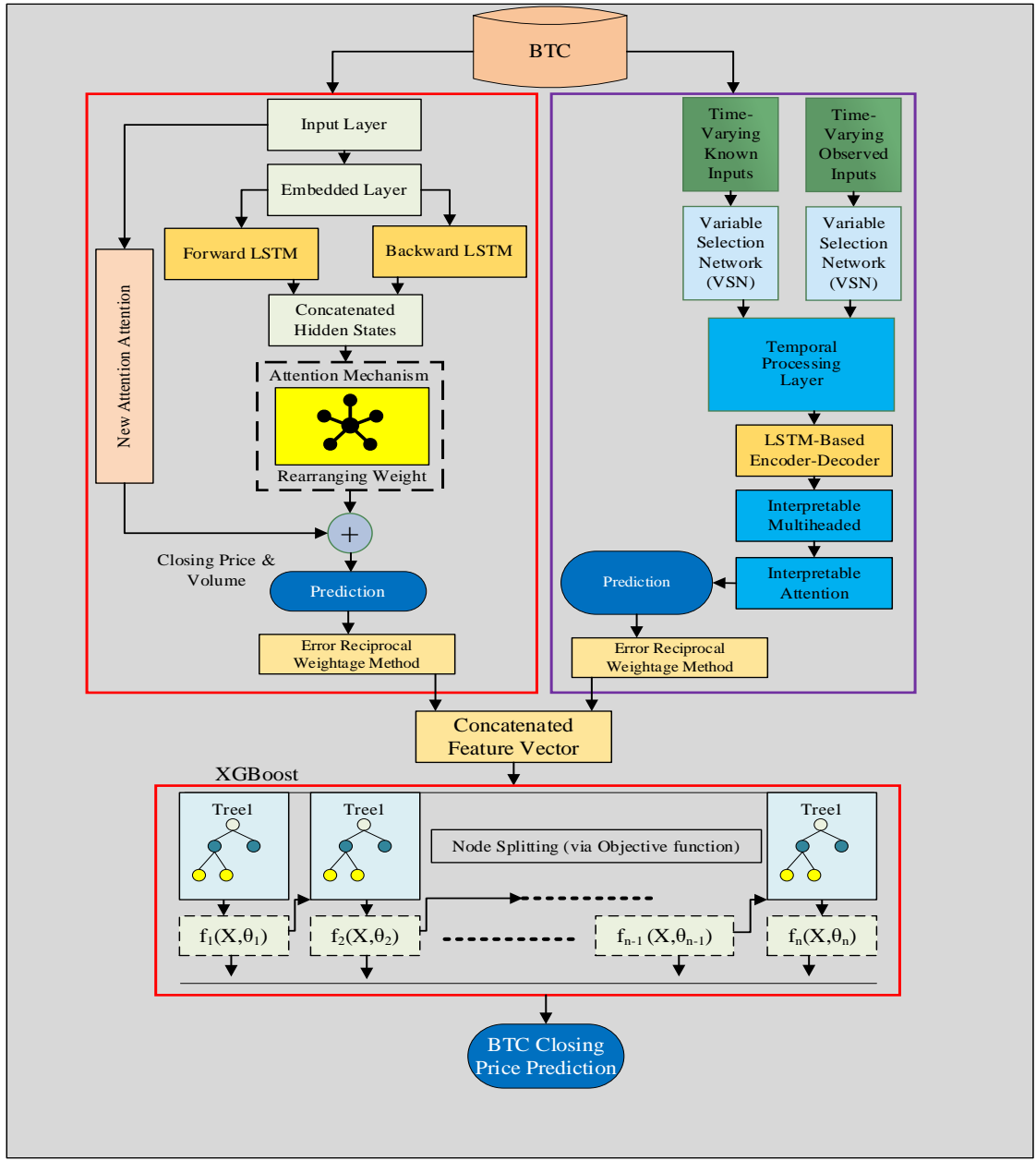

*Figure 3. Proposed TFT-ACB-XML detailed framework.*

## 3.1 TFT Model

TFT is designed for multivariate time series forecasting and is often used with variable selection and attention mechanisms that can aid interpretability. A customized TFT is used in this study to balance model complexity with forecast efficiency. Since static covariates are absent from the dataset, the customized TFT focuses on the temporal dynamics of BTC market features using daily OHLCV variables. Standard multi-quantile heads are omitted to reduce complexity since the study targets point forecasts. This design maintains compatibility with the subsequent XGBoost meta-learner, which refines residual errors based on point estimates. The customized TFT model as shown in Figure 4 begins with temporal variable selection networks. These networks assign importance weights to each input feature at each time step and suppress noisy contributions. Gated Residual Networks (GRNs) compute relevance scores and capture non-linear feature interactions. These scores are normalized through a softmax layer to determine each feature contribution. Gated Linear Units (GLUs) regulate information flow and improve training stability. The temporal encoder uses an LSTM to process sequences within the look-back window. A temporal self-attention layer assigns weights across the encoded sequence to highlight influential periods and support learning of long-range dependencies across the look-back window. A position-wise feedforward layer refines representations at each time step through non-linear projections and prepares the sequence for prediction.

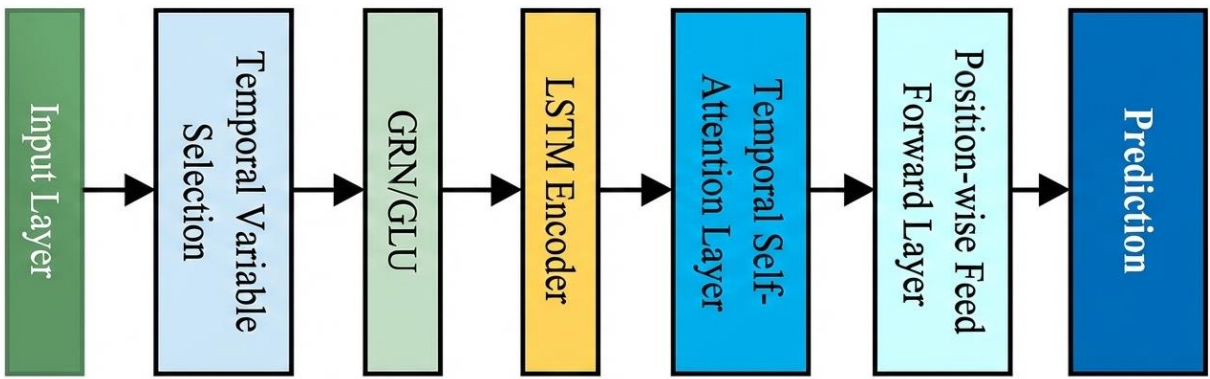

Figure 4. Temporal Fusion Transformer Architecture.

## 3.2 ACB Model

The TFT-ACB-XML framework uses ACB model to capture short-term sequential dependencies and local volatility. As illustrated in Figure 5 the BiLSTM extends the standard LSTM by incorporating both forward and backward passes over the historical look-back window. This bidirectional processing improves representation learning within the input window and does not use future observations beyond the forecast origin.

### 3.2.1 Customized Attention Mechanism

Two attention strategies are embedded within the Customized BiLSTM to focus more on high impact market features. First, an attention gate substitutes the conventional forget gate by attending solely to historical cell states and independent of the current input signal. This modification simplifies the architecture and reduces the number of trainable parameters. Figure 6 shows the conceptual difference between the standard and customized designs. The state update logic for the customized cell is defined as follows:

$$A_t = \sigma (W_a \cdot [C_{t-1}] + b_a) \quad (1)$$

$$C_t = A_t \odot C_{t-1} + i_t \odot \zeta_t \quad (2)$$

where $A_t$ represents the Attention Gate output and replaces the standard forget gate $f_t$.

Second, a new feature level attention weighting mechanism is used in parallel to the BiLSTM to give high priority to daily closing price and volume to enhance responsiveness to liquidity-driven fluctuations. The customized attention block is shown in Figure 7.

### 3.3 Weighting Strategy

An error-based inverse method is used to optimize weight allocation within the TFT-ACB-XML

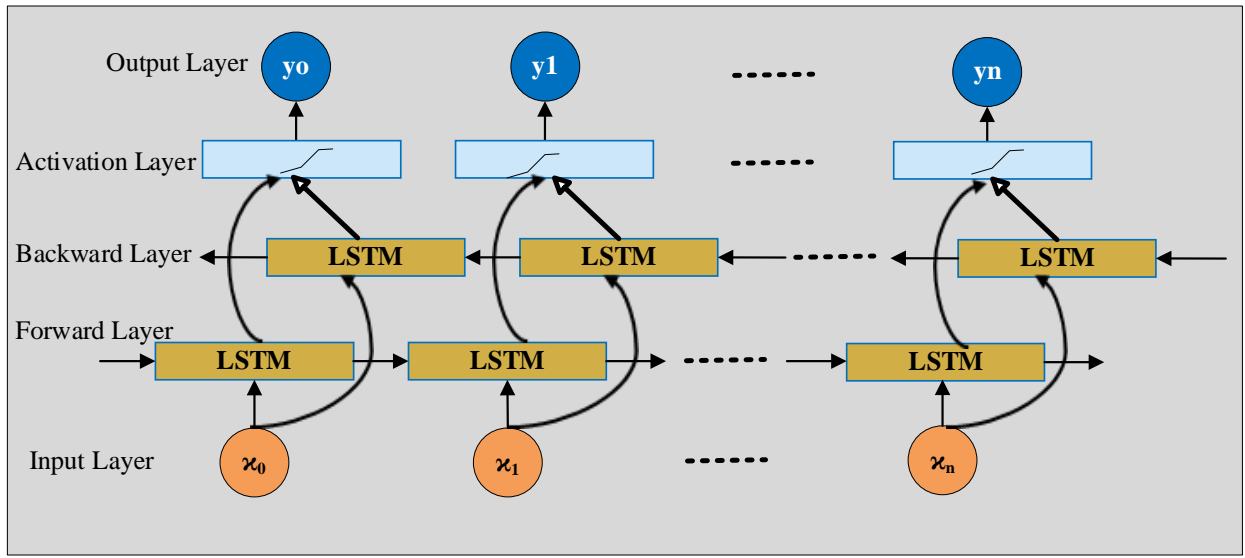

*Figure 5. BiLSTM structure.*

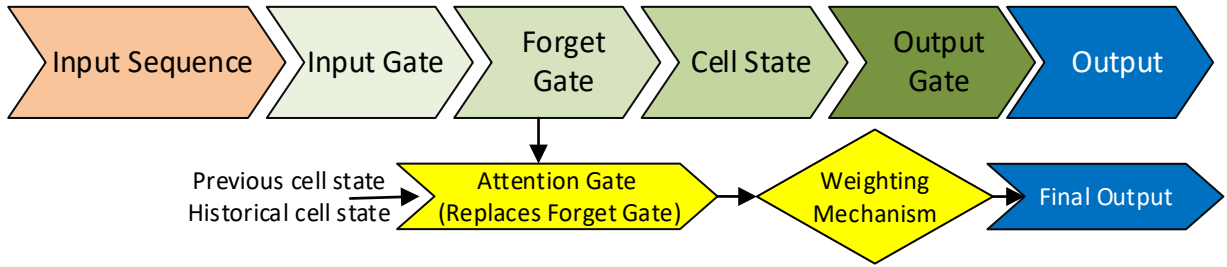

*Figure 6. Traditional LSTM vs BiLSTM with attention mechanism.*

framework and thereby improve predictive accuracy. This strategy systematically attenuates the influence of models that yield higher forecasting errors and gives greater influence on models with lower prediction errors by assigning weights inversely proportional to their historical performance. Weights are calculated using MAPE (Mean Absolute Percentage Error) as the primary assessment criterion. The mathematical formulation is as follows:

$$P_{meta}(t) = [W_{bl} \times P_{bl(t)}, W_{tft} \times P_{tft(t)}], t = 1, 2, \ldots, n \quad (3)$$

$$W_{bl} = \frac{1/Error_{bl}}{1/Error_{bl} + 1/Error_{tft}} \quad (4)$$

$$W_{tft} = \frac{1/Error_{tft}}{1/Error_{bl} + 1/Error_{tft}} \quad (5)$$

In these expressions, $P_{meta}(t)$ represents a two-dimensional feature vector formed by concatenating the weighted one-step-ahead predictions of the ACB and customized TFT model, while $W_{bl}$ and $W_{tft}$ are scalar inverse-error weights computed once on the validation split using MAPE. During the walk-forward evaluation on the test split, base learner parameters, $W_{bl}$, $W_{tft}$ and the trained XGBoost model are kept fixed. The terms $Error_{bl}$ and $Error_{tft}$ represent validation MAPE of ACB and customized TFT and $P_{bl}(t)$, and $P_{tft}(t)$ denote their unweighted outputs. This meta-feature vector is provided as input to the XGBoost meta-learner to capture non-linear residuals and correct individual model biases for the final prediction.

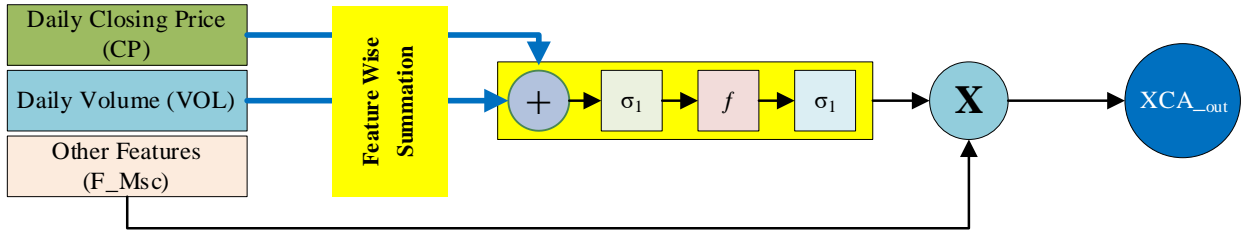

*Figure 7. Structure of new customized-attention mechanism.*

### 3.4 XGBoost

XGBoost is the final refinement layer of TFT-ACB-XML and focuses on the weighted predictions produced by the Customized TFT and ACB models. XGBoost is utilized as an optimized gradient-boosted decision tree ensemble to capture the non-linear residuals and error patterns that customized TFT and ACB models may overlook. The algorithm refines the predictions from the customized TFT and ACB models by incrementally combining weak learners and allows it to adapt to shifting dynamics and maintain stable accuracy. The XGBoost meta-learner helps the proposed framework to adapt to the varying market conditions and improve accuracy.

## 3.5 Evaluation Metrics

Three standard error metrics i.e. MAPE, MAE (Mean Absolute Error) and RMSE (Root Mean Squared Error) are used to evaluate predictive accuracy. MAPE is used to assess relative accuracy, MAE measures absolute deviation and RMSE penalizes larger prediction errors. These are computed as:

$$X_{MAPE} = \frac{1}{n}\sum_{i=1}^{n} \frac{|P_{true}-P_{fcst}|}{P_{true}} \times 100 \quad (6)$$

$$X_{MAE} = \frac{1}{n}\sum_{i=1}^{n} |P_{true} - P_{fcst}| \quad (7)$$

$$X_{RMSE} = \sqrt{\frac{1}{n}\sum_{i=1}^{n} (P_{true} - P_{fcst})^2} \quad (8)$$

Variable $P_{true}$ and $P_{fcst}$ represent the actual BTC price and corresponding forecast at instance *i*. The total number of observations is denoted by *n*. These metrics capture the predictive accuracy and error sensitivity to provide a robust evaluation of forecasting performance across the volatile institutionalization era.

## 3.6 Models Training and Testing

The proposed framework uses a multi-stage approach for BTC price prediction. The training and testing process consists of four sequential phases: data preparation, parallel model execution, weighted fusion and final-stage refinement. Figure 8 illustrates the structure of the presented framework.

**Stage 1: Data Preprocessing**: Stage 1 begins with feature selection and input variables include daily open, high, low, close prices and trading volume. Features are normalized using Min-Max scaling fitted on the training split only, then applied to validation and test splits using the same scaling parameters. The dataset is partitioned chronologically into training (80% - 01/10/2014 - 05/10/2023), validation (10% 06/10/2023 - 19/11/2024) and testing (10% 20/11/2024 - 05/01/2026) to avoid look-ahead bias.

**Stage 2: Parallel Models Prediction:** The customized TFT module is trained on multi-year historical data but uses a fixed input context (L = 60 days), which supports generalization across different market regimes and changing feature relevance. The ACB module uses its customized BiLSTM and attention mechanism to emphasize short-term price movements and their association with changes in trading volume patterns. This stage produces two independent forecasts: one from

TFT ($P_{tft}$) and one from ACB ($P_{bl}$) and uses the same look-back window (L = 60 days) for both base learners.

**Stage 3: Weighted Stacking and XGBoost Refinement:** In this stage, inverse-error weights $W_{tft}$ and $W_{bl}$ are computed using MAPE on the validation split, using validation forecasts produced by base learners trained on the training split. The weights are applied to the corresponding base predictions to form the weighted meta-feature vector $P_{meta}(t) = [W_{bl} \times P_{bl(t)}, W_{tft} \times P_{tft(t)}]$. The weighted predictions from TFT and ACB are concatenated as a two-dimensional meta-feature vector and used to train the XGBoost regressor. The test split is not used during weighting or meta-learner training. The meta-learner improves the final forecast by learning how to combine both signals.

**Algorithm 1: Time-series stacking protocol**

I. Split the series chronologically into train (80%), validation (10%), test (10%).
i. Train the base learners (customized TFT and ACB) on the train split.
ii. Generate one-step-ahead predictions on the validation split using the trained base learners.
iii. Compute inverse-error weights $W_{tft}$ and $W_{bl}$ on the validation split using MAPE.
iv. Form weighted meta-features on the validation split as $P_{meta}(t) = [W_{bl} \times P_{bl(t)}, W_{tft} \times P_{tft(t)}]$.
v. Train XGBoost on ($P_{meta}(t), P_{true}(t)$) using the validation targets.
vi. Test (held-out): generate walk-forward predictions over the test split using base learners trained on the train split. Apply the fixed scalar weights $W_{tft}$ and $W_{bl}$ (computed on the validation split) and obtain final forecasts using the fixed trained XGBoost meta-learner.
vii. The test split is used only for final evaluation and is never used for training, weighting, or hyperparameter selection.

**Stage 4: Evaluating Prediction Performance:** Stage 4 evaluates one-step-ahead prediction performance with MAPE, RMSE and MAE on the test block under the walk-forward procedure.

## 4. Experimental Setup

This experimental setup section describes the implementation and validation of the TFT-ACB-XML framework. The experiments evaluate accuracy and robustness under abrupt BTC market shifts.

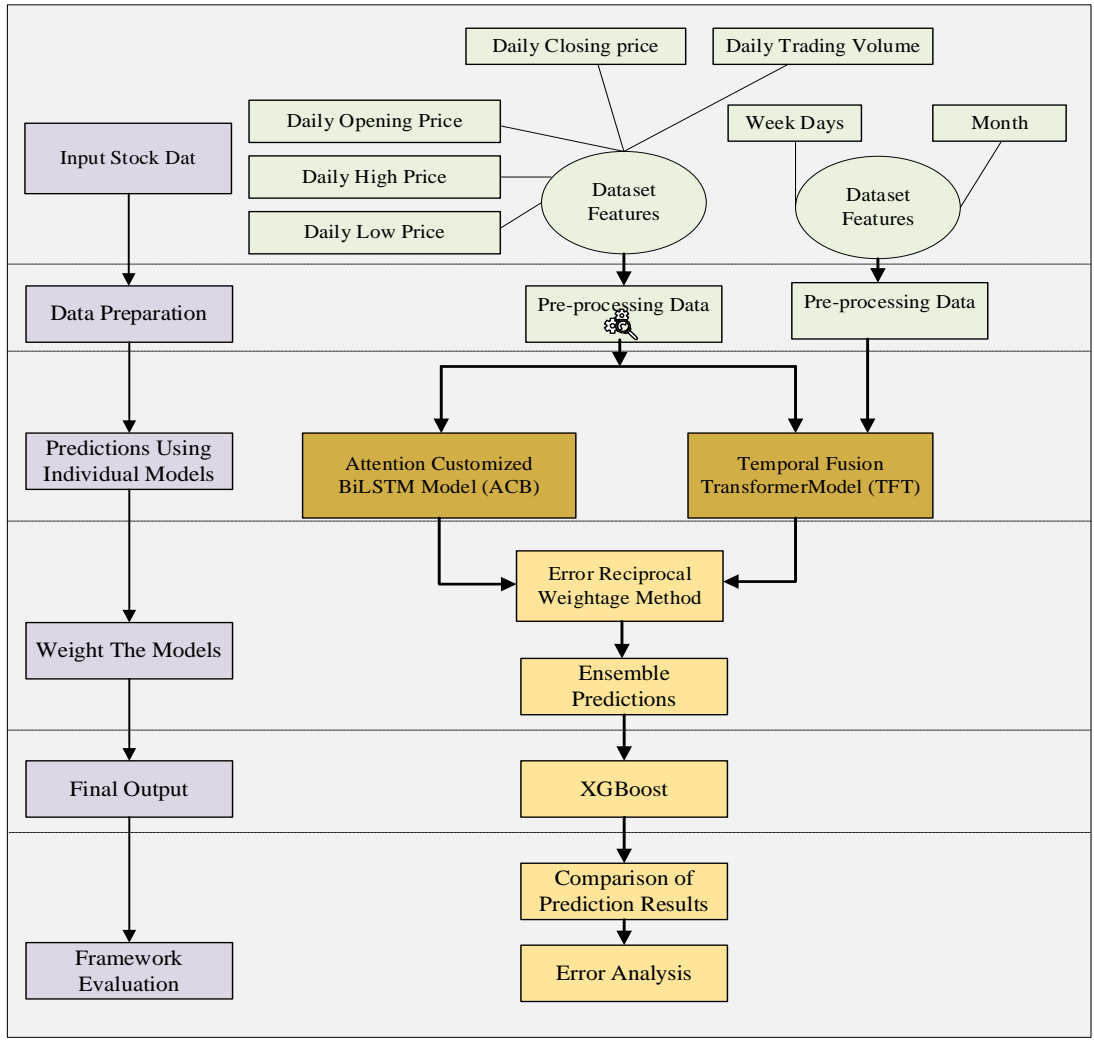

*Figure 8. BTC price prediction methodology in the proposed TFT-ACB-XML framework.*

## 4.1 Evaluation Methodology

The proposed framework is implemented using Python to evaluate predictive performance. Experiments compare the proposed framework with its individual components and other baseline models, as reported in Table 1. The experimental design reduces evaluation bias by using the same chronological train/validation/test split for all models and the same 60-day look-back. For neural baselines, hyperparameters are selected using the validation split only. For stacked models, the meta-learner is trained on the temporally held-out validation block to combine base-model forecasts without using any information from the test period, ensuring a leakage-free stacking procedure for time series. The XGBoost configuration is reported in

*Table 9* and uses regularization and early stopping to limit overfitting. The test split is used exclusively for final evaluation and is never used for training, weighting or hyperparameter selection.

### 4.2 Data Acquisition and Preprocessing

Data acquisition and preprocessing play a central role in BTC price forecasting. The accuracy of the framework depends on input data quality and preprocessing. This study implements a multi-stage data management process that consists of acquisition, cleaning and normalization of data to ensure that the proposed hybrid framework captures linear as well as non-linear market patterns. Preprocessing transforms raw historical market signals into a refined format and mitigates the impact of noise and extreme outliers common within digital asset markets.

#### 4.2.1 Data Source

Assessment of available data sources showed that Yahoo Finance provides a convenient public source for daily OHLCV data used in this study. The dataset includes daily features such as date, opening and closing prices, high and low prices and trading volume. This study uses a dataset obtained from Yahoo Finance covering the period from October 1, 2014 to January 5, 2026.

**Feature Engineering (Calendar Covariates)**

Before model training, two calendar features, day-of-week and month, were generated from the Date column using a simple preprocessing script. These variables are deterministic and known in advance for each timestamp. They were one-hot encoded and included only as additional known-ahead covariates for the TFT model, while the ACB model used the OHLCV variables only. The calendar features were derived directly from the dataset Date field without time zone conversion.

**Data Availability:**

The following Yahoo Finance link provides access to the dataset.

https://finance.yahoo.com/quote/BTC/history/?period1=1412121600&period2=1767657600

The BTC-USD daily OHLCV dataset is publicly available from Yahoo Finance. The processed dataset used in this study is available at the GitHub at the following link:

https://raw.githubusercontent.com/itsriaz/BTC/refs/heads/main/TFT_%20Paper_DataSet.csv

*Table 1.Evaluation Models*

| Category | Baseline Models |
|---|---|

| Traditional & Machine Learning | Naïve persistence baseline, XGBoost |
|---|---|
| Sequential Deep Learning | RNN, LSTM, GRU, BiLSTM, Stacked LSTM |
| Hybrid & Attention Models | Attention LSTM, CNN LSTM, Attention BiLSTM, ACB, ACB-XDE |
| State of the Art Transformers | Informer, iTransformer, TFT |
| Proposed Framework | TFT-ACB-XML |

### 4.2.2 Data Cleaning and Outlier detection

The dataset obtained from Yahoo Finance was reviewed for accuracy and completeness. As no missing values were identified, there was no need for data imputation. In addition, outlier detection based on the Z-score method (threshold set at ±3) was applied to isolate real price movements from potential data anomalies. In the dataset of 4115 entries, 62 data points are identified as outliers and constitute approximately 1.51% of the dataset. Z-scores were computed using training-split statistics and then applied to the full series for reporting. These points were cross-checked against the original OHLCV records. A detailed review of these cases confirmed that the identified values reflect actual price volatility rather than recording errors or noise. Therefore, these outliers were retained to preserve the true market behavior within the dataset. The first five outliers exceeding ±3 are shown in the Table 2 and Figure 9 shows the Z-score analysis for outlier detection.

Table 2. Outliers Z score greater than 3.

| Date | Open | High | Low | Volume | Close |
|---|---|---|---|---|---|
| 11/01/2021 | 38347.0 | 38347.0 | 30550.0 | 1.23E+11 | 35567.0 |
| 29/01/2021 | 34319.0 | 38406.0 | 32065.0 | 1.18E+11 | 34316.0 |
| 08/02/2021 | 38887.0 | 46204.0 | 38076.0 | 1.01E+11 | 46196.0 |
| 09/02/2021 | 46185.0 | 48004.0 | 45167.0 | 9.18E+10 | 46481.0 |
| 22/0 | | | | | |

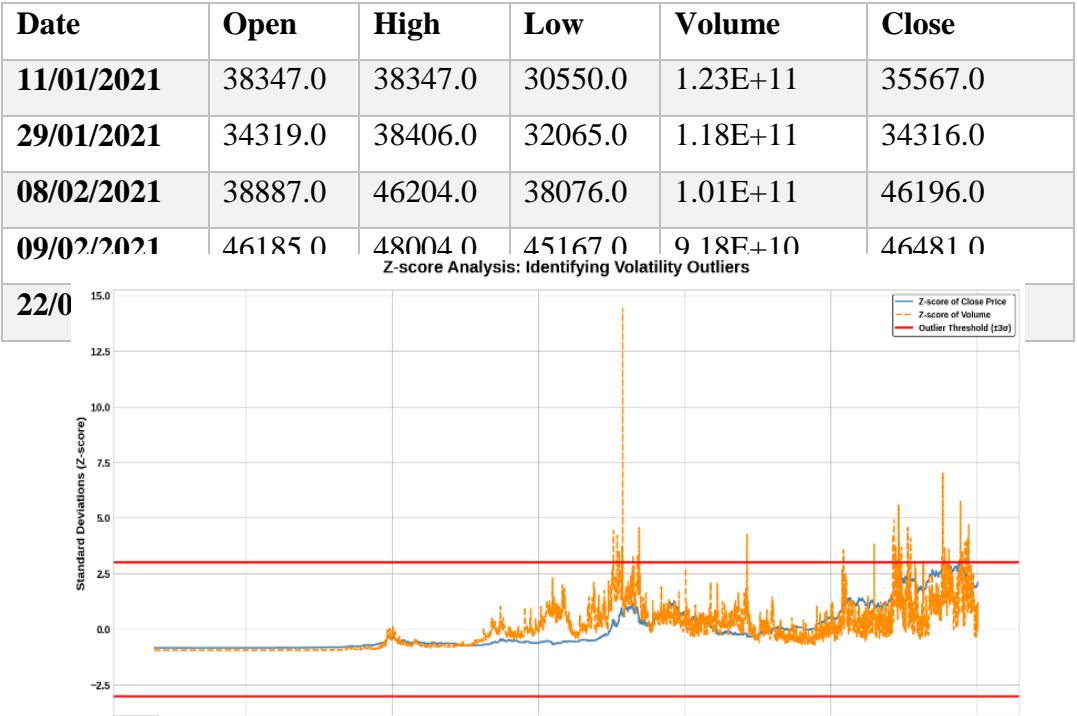

Figure 9. Z-score Analysis for Outlier Detection.

### 4.2.3 Data Splitting and Normalization

For experimental evaluation, the BTC dataset was used to support robust model training and unbiased performance assessment. The dataset contains daily OHLCV data from October 1, 2014, to January 5, 2026. The series was sorted chronologically and split into training (80%), validation (10%) and test (10%) sets. For TFT, two calendar covariates (day-of-week and month) are additionally derived from the date index and used as known-ahead inputs, while the ACB model uses OHLCV variables only. The validation split is used to compute error-reciprocal weights for stacking the base learners before meta-learner refinement. The most recent portion of the series is reserved for testing to ensure that no future information enters training or model selection.

Daily OHLCV inputs are normalized using Min–Max scaling with parameters fitted on the training split only. The fitted training min and max values are then used to transform the validation and test splits using the same parameters. No clipping is applied after transformation; therefore validation/test values may fall outside the [0,1] range when they exceed the training range. It avoids saturation effects in trending series and preserves information about new extremes. This behavior is reflected in the post-normalization summary statistics, where some features exceed 1.0 during later periods. **Error! Not a valid bookmark self-reference.** shows the dataset structure before and after normalization, Table 4 reports summary statistics and Figure 10 provides a visual illustration of the normalization effect. For the TFT model, two additional calendar covariates (day-of-week and month) are derived from the date index. These variables are used as known-ahead inputs and are one-hot encoded. The one-hot calendar features are kept as binary indicators (0/1) and are not Min–Max scaled in the same way as OHLCV.

*Table 3. Structure of Data.*

| Dataset Before Normalization | | | | | | |
|---|---|---|---|---|---|---|
| **Date** | **Split** | **Open** | **High** | **Low** | **Volume** | **Close** |
| **01/10/2014** | train | 387.43 | 391.38 | 380.78 | 26229400.00 | 383.61 |
| **02/10/2014** | train | 383.99 | 385.50 | 372.95 | 21777700.00 | 375.07 |
| **03/10/2014** | train | 375.18 | 377.70 | 357.86 | 30901200.00 | 359.51 |
| **04/10/2014** | train | 359.89 | 364.49 | 325.89 | 47236500.00 | 328.87 |
| **05/10/2014** | train | 328.92 | 341.80 | 289.30 | 83308096.00 | 320.51 |
| | | | | | | |
| Dataset After Normalization | | | | | | |
| **Date** | **Split** | **Open minmax** | **High minmax** | **Low minmax** | **Volume minmax** | **Close minmax** |
| **01/10/2014** | train | 0.00312485 | 0.00261962 | 0.003160674 | 5.7878E-05 | 0.00304965 |

| 02/10/2014 | train | 0.003073806 | 0.002533849 | 0.003042355 | 4.51949E-05 | 0.002922878 |
| 03/10/2014 | train | 0.002943085 | 0.00242008 | 0.002814491 | 7.11882E-05 | 0.002691978 |
| 04/10/2014 | train | 0.002716154 | 0.002227482 | 0.002331592 | 0.000117728 | 0.002237214 |
| 05/10/2014 | train | 0.002256384 | 0.001896675 | 0.001778961 | 0.000220498 | 0.002113217 |

*Table 4. Summary of data before and after pre-processing.*

| | Before Pre-processing/Normalization | | | | |
|---|---|---|---|---|---|
| | Open | High | Low | Volume | Close |
| count | 4115.00 | 4115.00 | 4115.00 | 4115.00 | 4115.00 |
| mean | 27004.77 | 27538.11 | 26422.38 | 21716122200.80 | 27005.87 |
| std | 31731.92 | 32255.68 | 31137.30 | 22880782068.72 | 31706.18 |
| min | 176.90 | 211.73 | 171.51 | 5914570.00 | 178.10 |
| 0.25 | 2785.08 | 2893.54 | 2688.23 | 1379424960.00 | 2805.18 |
| 0.5 | 10821.63 | 11095.87 | 10528.89 | 17137029730.00 | 10844.64 |
| 0.75 | 42323.23 | 43184.64 | 41526.62 | 33237193369.00 | 42367.22 |
| max | 124752.14 | 126198.07 | 123196.05 | 351000000000.00 | 124752.53 |
| | After Pre-processing/Normalization (No clipping, train-fit MinMax) | | | | |
| | Open | High | Low | Volume | Close |
| count | 4115.00 | 4115.00 | 4115.00 | 4115.00 | 4115.00 |
| mean | 0.40 | 0.40 | 0.40 | 0.06 | 0.40 |
| std | 0.47 | 0.47 | 0.47 | 0.07 | 0.47 |
| min | 0.00 | 0.00 | 0.00 | 0.00 | 0.00 |
| 0.25 | 0.04 | 0.04 | 0.04 | 0.00 | 0.04 |
| 0.5 | 0.16 | 0.16 | 0.16 | 0.05 | 0.16 |
| 0.75 | 0.63 | 0.63 | 0.62 | 0.09 | 0.63 |
| max | 1.85 | 1.84 | 1.86 | 1.00 | 1.85 |

This study uses walk-forward evaluation on the held-out test block. After each one-step-ahead prediction, the next day's observed OHLCV values are appended to the input history for the subsequent forecast. During the test window, model hyperparameters and stacking weights remain fixed. The input window is updated with new observations, but model weights are not re-estimated during the test block.

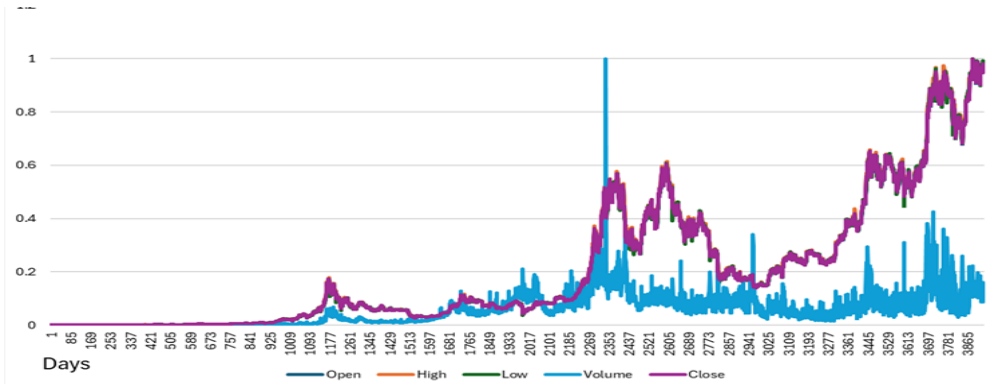

*Figure 10. Normalized Features Over Time*

### 4.2.4 Regime Context Around 2024 Market Events

BTC forecasting became harder due to the 2024 spot ETFs and the 2024 block halving and the price series shows clear changes in volatility and trading activity in that part of the timeline. Descriptive statistics are computed on daily BTC OHLCV data from October 1, 2014 to January 5, 2026 to support this regime-shift motivation using the same data analyzed in this study. The analysis uses three windows:

i. 180 days before January 10, 2024 (Pre-ETF)
ii. January 10, 2024 to April 19 2024 (ETF→ Halving)
iii. 180 days after April 20, 2024 (Post-halving).

Daily log returns are defined as $r_t = \ln(Close_t / Close_{t-1})$. Volatility is reported in annualized form as the standard deviation of log returns scaled by $\sqrt{365}$ and the rolling volatility curve uses a 30-day window. Table 5 reports shift in return dispersion and volume levels across the three windows, and Figure 11 shows the rolling volatility and volume dynamics. These results indicate that the post-2024 segment has different statistical behavior than earlier periods in this dataset and supports treating it as a more challenging forecasting regime for evaluation.

*Table 5. Descriptive regime context around 2024 market events.*

| Regime | Date range | N (days) | Mean log return (%) | Std log return (%) | Annualized volatility | Median Volume | Mean Volume | Median Close (USD) |
|---|---|---|---|---|---|---|---|---|
| **Pre-ETF (−180d)** | 2023-07-14 → 2024-01-09 | 180 | 0.212 | 2.106 | 0.402 | 1.43e10 | 1.69e10 | 29,781.91 |
| **ETF→ Halving** | 2024-01-10 → 2024-04-19 | 101 | 0.322 | 3.159 | 0.604 | 3.37e10 | 3.51e10 | 61,276.69 |
| **Post-halving (+180d)** | 2024-04-20 → 2024-10-17 | 181 | 0.032 | 2.563 | 0.490 | 2.90e10 | 2.93e10 | 63,049.96 |

### 4.3 Models Training and Evaluation

TFT-ACB-XML framework's individual models undergo structured training and testing. The ACB and the TFT models are trained independently on 80% of the daily sampled historical BTC data. A validation split of 10% estimates error-based weights and the final 10% is held out as an unseen test block for accuracy assessment. Predictions from both customized TFT and ACB are first weighted through an error-reciprocal weighting strategy. The weights are derived from validation performance to prevent data leakage and models with lower prediction errors are assigned higher weight. Finally, these weighted outputs are then concatenated into a feature vector and fed to an XGBoost regressor. XGBoost captures non-linear residuals and produces the final BTC closing price prediction. The final output of the proposed framework is assessed against the actual value of BTC. Testing uses a walk-forward evaluation on the test block with fixed model parameters and fixed weights and no refitting occurs on test targets. The same walk-forward evaluation protocol and the same test block are applied to all baseline models to ensure a fair comparison.

### 4.4 Evaluation Criteria

Evaluation criteria rely on standard statistical measures widely used in time-series forecasting to assess the accuracy of the proposed TFT-ACB-XML framework. Three error metrics are used for evaluation: MAPE, RMSE and MAE. The primary indicator for this study is MAPE and measures average percentage deviation between forecasted value and the actual BTC prices.

$$MAPE = \frac{100}{n} \sum_{i=1}^{n} \frac{|P_{true} - P_{fcst}|}{P_{true}} \qquad (9)$$

Analyses of MAPE are complemented by the other two metrics such as RMSE and MAE. RMSE assigns more penalty to large errors to identify unstable predictions. MAE gives the mean absolute difference and is used as a reference for overall accuracy.

$$MAE = \frac{1}{n}\sum_{i=1}^{n}|P_{true} - P_{fcst}| \qquad (10)$$

$$RMSE = \sqrt{\frac{1}{n}\sum_{i=1}^{n}(P_{true} - P_{fcst})^2} \qquad (11)$$

Within in the above equations (10) and (11), n represents the number of BTC data, $p_{true}$ is the actual BTC price while $p_{fcst}$ the forecasted BTC price. Collectively, these metrics provide a balanced assessment of predictive performance of the proposed framework.

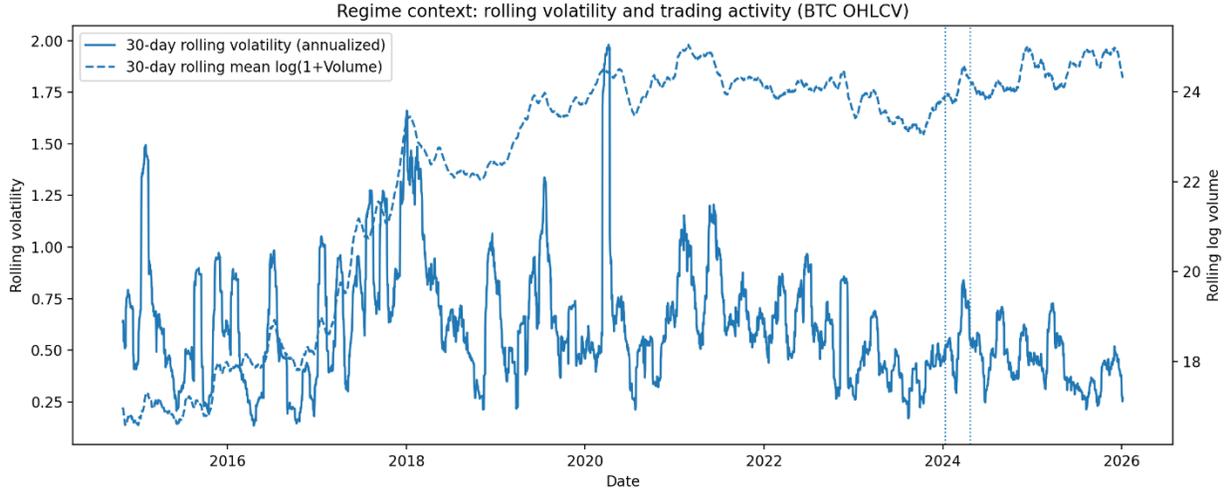

*Figure 11. Regime context around 2024 market events.*

Confidence interval analysis further examines the robustness of the proposed framework using MAPE as the reference metric. A 95% confidence level (CI) [42] estimates variability in predictive accuracy through the standard formulation shown in the following Equation 12

$$Confidence\ interval\ (CI) = MAPE \pm Z_{\alpha/2} \times \frac{\sigma MAPE}{\sqrt{n}} \qquad (12)$$

$Z_{\alpha/2}$ is the Z-score representing the standard score linked to the selected confidence level for a 95% confidence interval $Z_{\alpha/2}$=1.96. $\sigma MAPE$ represents dispersion across prediction errors. $n$ denotes the total number of evaluation samples.

CI is reported as an approximate summary of variability and is complemented by repeated runs. This analysis measures the average error and also the consistency of the proposed framework. Further, stability is assessed using ten independent trials with different random seeds, where TFT, ACB and the XGBoost meta-learner are re-trained from scratch under the same chronological split and fixed hyperparameter settings.

### 4.5 Hardware setup

The experimental environment is based on a laptop equipped with an AMD Ryzen 7 4800H processor having 8 cores / 16 threads operating at 3 GHz and 32 GB of RAM. The proposed framework is implemented using Python. TensorFlow and Keras are employed for developing the ACB and customized TFT models and the XGBoost component is implemented using the XGBoost Python library. Standard numerical and data handling libraries such as NumPy, Pandas and scikit-learn are used for preprocessing and evaluation tasks.

### 4.6 Parametric Configurations

This section describes the parameter settings used to fine-tune deep learning and ensemble components of the TFT-ACB-XML framework. These components are the TFT, ACB and XGBoost. These parameters influence forecasting accuracy, generalization and computational complexity. Clear documentation of these configurations helps in reproducibility and facilitates performance optimization on comparable forecasting tasks.

#### 4.6.1 TFT Parameter Configurations

The customized TFT model parameters are selected to support one-step-ahead forecasting and preserve interpretability. Important parameters include the number of attention heads, hidden layer dimensions, dropout rates and the gating components. These settings control the model's ability to learn temporal hierarchies and adapt to feature relevance over time. Table 6 and
Table 7 show the final configuration and parameter count that ensure a balance between predictive accuracy and computational cost and making it suitable for high-variance financial data like BTC prices.

In this study, the customized TFT model uses a sliding input window of 60 days with five OHLCV features same as the ACB input. In addition, two calendar covariates derived from the date index (day-of-week and month) are included for TFT as known-ahead inputs and are one-hot encoded. Inputs are first processed by a temporal variable selection network with GRNs and GLUs to filter noisy features and emphasize relevant patterns. The resulting representations are fed into a 32-unit LSTM encoder with a dropout rate of 0.1 to reduce overfitting. A single-head temporal self-attention layer follows the encoder to assign attention weights across time steps and highlight influential periods during the post-2024 regime (including the spot BTC ETFs period). A position-wise feedforward layer refines representations at each time step and a final dense output layer produces point forecasts for BTC prices. The customized configuration contains approximately

15,457 trainable parameters (for d_model = 32) and provides sufficient learning capacity while keeping computational cost low. The customized TFT model is trained using a learning rate of 0.001 with MSE as the loss function. Gradient clipping at 0.1 is applied to handle the extreme volatility of the BTC market.

*Table 6. Parameter settings of TFT.*

| Parameter | Value | Description |
| --- | --- | --- |
| Encoder Length | 60 | Past time steps (context window) |
| Decoder Length | 1 | Future time steps (prediction horizon) |
| Learning Rate | 0.001 | Step size for optimizer |
| Loss Function | MSE | Optimized objective |
| Gradient Clipping | 0.1 | Maximum gradient norm |
| Variable Selection GRN | 4,224 | 4×(d_model×d_model+d_model) |
| LSTM Encoder | 8,320 | 4×(d_model×(d_model+input_dim)+d_model) |
| Temporal Self-Attention | 1 Head-4,224 | (d_model×d_model×3+d_model×d_model)×n_head |
| Position-wise Feed-Forward | 2,112 | 2×(d_model×d_model+d_model) |
| Output Layer (Dense) | 33 | d_model+1 |
| Dropout Rate | 0.1 | Regularization probability |
| Batch Size | 32 | Samples per batch |
| Total Parameters | 15,457 | Sum of customized components |

*Table 7. TFT Parameter Counts.*

| Layer / Component | Formula (per unit) | Customized (d_model=32) | Full (d_model=64) |
| --- | --- | --- | --- |
| **Variable Selection GRN (encoder/decoder)** | 4 × (d_model × d_model + d_model) | 4,224 | 16,640 |
| **LSTM Encoder** | 4 × (d_model × (d_model + input_dim) + d_model) | 8,320 | 17,920 |
| **Temporal self-attention** | (d_model × d_model × 3 + d_model × d_model) × n_head | 4,224 | 16,640 |

| Position-wise Feed-Forward | 2 × (d_model × d_model + d_model) | 2,112 | 8,320 |
| Output Layer (Dense) | d_model + 1 | 33 | 65 |
| Total Parameters | — | ≈ 15,457 | ≈ 59,585 |

### 4.6.2 ACB Parameter Configurations

The predictive accuracy of the ACB model depends on its architectural design and training configuration. A look-back window of 60-time steps is used to handle the non-linear dynamics and short-term volatility associated with the BTC market during the institutionalization phase. Training is performed using the Adam optimizer with an initial learning rate of 0.001 and MSE as the loss function to maintain stable gradient updates during the parallel execution phase. The ACB integrates a feature-attention branch with a two-layer BiLSTM encoder. The attention branch emphasizes daily closing price and trading volume to generate a context vector that highlights key market signals. In parallel, the BiLSTM layers (comprising 64 and 32 units respectively) process the full OHLCV sequence to capture temporal dependencies. Their outputs are concatenated and passed to a dense regression head. A dropout rate of 0.2 and early stopping with a patience of 10 epochs are implemented to prevent overfitting. Based on these settings, the baseline configuration contains 80,926 trainable parameters. A reduced ablation version, which halves the hidden units to 32 and 16, contains 16,610 parameters to evaluate the performance-complexity trade-offs. *Table 8* shows the detailed parametric configuration of the ACB model.

### 4.6.3 XGBoost Parameter Configuration

The assessment and performance of the XGBoost model within the proposed TFT-ACB-XML framework are influenced by a number of parameters. These include the number of decision trees, the objective function, controls for model complexity, regularization terms, the subsampling rate for data, the choice of a weak learner and parameters controlling the iterative decision tree boosting process. The XGBoost model is configured with 1000 estimators, a maximum depth of 4, and a learning rate (eta) of 0.05. Table 9 shows the configuration of XGBoost parameter settings. The deep learning components provide high-level feature extraction and the gradient boosting trees provide a structured residual-correction stage and support the robustness of the proposed framework. The objective function is set to squared error (objective = reg:squarederror) to ensure training stability and the XGBoost minimizes squared error during the gradient boosting process.

The proposed framework's primary evaluation metric is MAPE and is used during the stacking phase to refine and assess the final price forecast. This distinction allows the model to leverage the mathematical efficiency of squared error optimization while ensuring the final output is robust according to percentage-based error standards.

*Table 8. Parameters settings of ACB*

| Parameter | Value | Description |
| --- | --- | --- |
| Time Steps | 60 | Historical window for temporal patterns |
| Input Features | 5 | Daily OHLCV signals |
| Optimizer | Adam | Adaptive moment estimation |
| Learning Rate | 0.001 | Gradient descent step size |
| Loss Function | MSE | Objective for price regression |
| Batch Size | 32 | Samples processed per gradient update |
| Input Projection Dim | 384 | Linear projection to higher dimension |
| Feature Attention Params | 65 | Focus on Closing-price and Volume |
| BiLSTM Layer 1 | 64 Units-49,536 | Bidirectional (2×24,768) |
| BiLSTM Layer 2 | 32 Units- 30912 | Bidirectional (2×15,456) |
| Dense Head | 64 → 1 | Final linear regression layer |
| Dropout Rate | 0.2 | Applied after BiLSTM layers |
| Early Stopping | 10 | Patience for validation loss |
| Baseline Total | 80,926 | Full architectural capacity |
| Reduced Total | 16,610 | Ablation model with halved units |

*Table 9. Parameter settings of XGBoost.*

| Parameter | Value | Description |
| --- | --- | --- |
| Objective | reg:squarederror | The loss function minimized during the training process. |
| n_estimators | 1000 | Number of decision trees |
| early_stopping_rounds | 50 | Prevents under/over-training |
| max_depth | 4 | Maximum depth |

| verbosity | 1 | Model training progression |
| --- | --- | --- |
| subsample | 0.7 | Fraction of samples per tree (1 = no subsampling |
| colsample_bytree | 0.7 | Regularization knob |
| learning_rate (eta) | 0.05 | Sequential construction of decision trees |
| alpha | 1 | Regularization parameters |
| reg_lambda | 5 | Stabilizes the model. |
| gamma | 1 | Overall model complexity |
| booster | Gbtree | Weak learner |
| eval_metric | mae | Early stopping monitored on the validation block |
| min_child_weight | 10 | Prevents splits on very small/noisy partitions and reduces overfitting. |

## 5. Results

The proposed TFT-ACB-XML framework's predictive performance is evaluated through a series of structured experiments using historical BTC-USD data from October 1, 2014 to January 5, 2026. The core inputs are daily OHLCV variables, while the TFT branch additionally receives one-hot encoded calendar covariates (day-of-week and month) as known-ahead features; ACB uses OHLCV only. First, the dataset is used to train and evaluate the customized TFT and ACB models separately. training and validation curves for the customized TFT and ACB base models are shown in Figure 12. validation loss follows training loss without clear divergence. Early stopping is applied (patience = 10 epochs) for the deep learning models, which indicates stable training behavior and no clear overfitting trend on the validation curves. After the computation of individual errors, an inverse error-based weighting strategy is applied to the outputs of the customized TFT and ACB models. The weighting process adjusts the contribution of each base learner based on validation error by assigning greater weight to the model with lower prediction error. This ensures that a model demonstrating smaller errors is assigned higher weight. MAPE is used as the primary measure for this weighting process and weights are calculated using Equation (4) and Equation (5). These weighted outputs are then stacked as independent features and fed into the XGBoost meta-learner. The proposed TFT-ACB-XML framework is evaluated on the held-out test block using a walk-forward procedure with fixed base learner parameters, fixed weights and a fixed XGBoost model. All key error metrics consistently demonstrate improvement when using the TFT-ACB-XML framework. MAPE shows a relative reduction of 14.47% compared to the ACB-XDE baseline (MAPE 0.76%). Similarly, MAE decreases by 4.92% (from 208.40 to 198.15) and RMSE decreases by 4.38% (from 270.14 to 258.30) as presented in Table 10. Furthermore, a benchmark is established using a naïve persistence baseline model that predicts the next-day closing price as identical to the previous day's price. This baseline produced a MAPE of 2.00%, an MAE of 559.89 USD and an RMSE of 1104.13 USD. In contrast, the proposed TFT-ACB-XML framework achieved a MAPE of 0.65% and shows an improvement of over 67.5% compared with the naïve persistence baseline.

Table 11 shows the PG of the proposed framework relative to naïve persistence baseline. The approximate CI analysis for MAPE further indicates improved forecasting accuracy as reported in Table 12. A relatively narrow CI range (0.65 ± 0.12% absolute) supports stable performance during an evaluation window that includes the 2024 BTC halving and the spot BTC ETFs period.

*Table 10. Error analysis of the proposed framework and evaluation with the base models.*

| Model | MAPE (%) | MAE ($) | RMSE ($) |
|---|---|---|---|
| Naïve persistence baseline (Baseline) | 2.00% | 559.89 | 1104.13 |
| RNN | 1.95 | 485.20 | 610.45 |
| LSTM | 1.78 | 442.50 | 560.12 |
| GRU | 1.72 | 428.10 | 545.30 |
| XGBoost | 1.62 | 402.40 | 512.18 |
| Attention-LSTM | 1.55 | 385.60 | 490.25 |
| BiLSTM | 1.48 | 368.15 | 472.60 |
| Stacked LSTM | 1.42 | 352.80 | 455.10 |
| CNN-LSTM | 1.35 | 335.40 | 432.75 |
| Attention- BiLSTM | 1.28 | 318.15 | 410.93 |
| Informer | 1.15 | 286.10 | 370.40 |
| iTransformer | 1.08 | 268.50 | 348.15 |
| Standard TFT | 1.02 | 253.40 | 330.20 |
| ACB | 0.82 | 242.60 | 292.80 |
| ACB-XDE | 0.76 | 208.40 | 270.14 |
| TFT-ACB-XML (Proposed) | **0.65** | **198.15** | **258.30** |

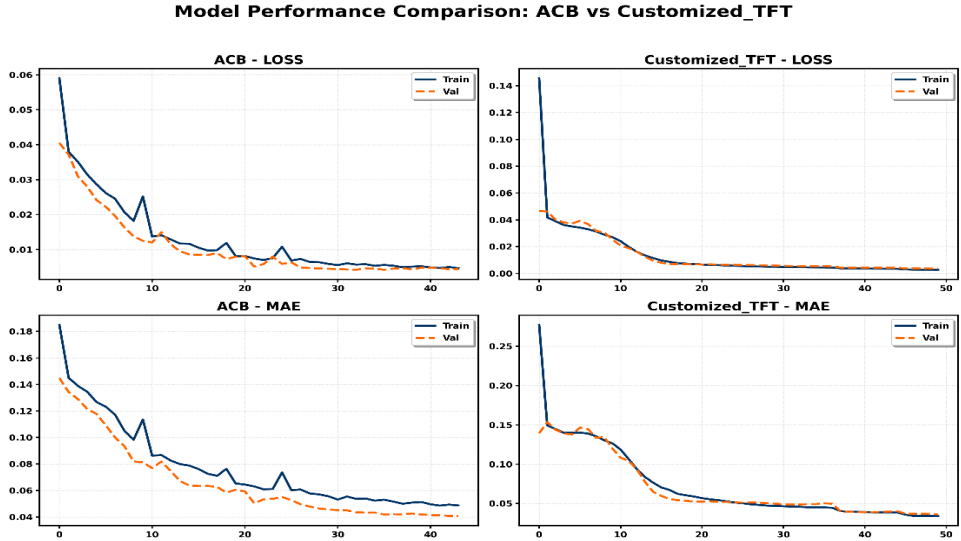

Figure 12. Training and validation performance of the ACB and customized TFT models

Table 11. PG of the proposed framework relative to naïve persistence baseline.

| | Model | MAPE | PG |
|---|---|---|---|
| 1 | TFT-ACB-XML (Proposed) | 0.65 | 0.675 |
| 2 | ACB-XDE | 0.76 | 0.62 |
| 3 | ACB | 0.82 | 0.59 |
| 4 | TFT | 1.02 | 0.49 |
| 5 | iTransformer | 1.08 | 0.46 |
| 6 | Informer | 1.15 | 0.425 |
| 7 | Attention-BiLSTM | 1.28 | 0.36 |
| 8 | CNN-LSTM | 1.35 | 0.325 |
| 9 | Stacked LSTM | 1.42 | 0.29 |
| 10 | BiLSTM | 1.48 | 0.26 |
| 11 | Attention-LSTM | 1.55 | 0.225 |
| 12 | XGBoost | 1.62 | 0.19 |
| 13 | GRU | 1.72 | 0.14 |

| 14 | LSTM | 1.78 | 0.11 |
| 15 | RNN | 1.95 | 0.025 |
| 16 | Naive Persistence | 2 | 0 |

## 5.1 Discussion

Prediction of BTC price is challenging due to extreme volatility, complex temporal dependencies, and non-stationary behavior in market data. Standalone BiLSTM models handle sequence dependencies well, but they can overfit on noisy financial data and may be less robust under regime changes. The proposed TFT-ACB-XML framework addresses these issues by incorporating complementary components that mitigate the limitations of BiLSTM and enhance prediction reliability. The customized TFT is trained on multi-year historical data using a fixed 60-day input context to handle long-term temporal dependencies and may offer qualitative insight and the ACB runs in parallel with customized TFT to capture short-term sequence dynamics efficiently. The customized TFT in this study uses a single LSTM encoder, single-head temporal self-attention, gating layers and point prediction outputs. It utilizes a variable selection network to filter relevant historical signals and regulates information flow to produce stable forecasts under volatile conditions.

As outlined in the framework methodology as shown in Figure 8, the independent predictions generated by both the customized TFT and ACB are first weighted through an error-reciprocal weighting strategy. This mechanism ensures that models with lower prediction errors are assigned higher weight to balance individual model strengths and improve overall stability. Finally, the individual predictions from the customized TFT and ACB are stacked as input features for the XGBoost meta-learner. XGBoost refines these stacked inputs by learning a non-linear combination of the base forecasts and reducing systematic error patterns. These three components collectively form a coherent and strategically layered architecture. Each component contributes distinct strengths: the customized TFT models temporal structure using gating, attention, and variable selection components, the ACB emphasizes short-term sequence dynamics, and XGBoost refines the combined predictions and ensures generalization. The proposed TFT-ACB-XML framework is empirically evaluated through extensive predictive experiments and demonstrated improved accuracy and resilience in BTC price forecasting.

*Table 12. MAPE Confidence interval variation.*

| Model | MAPE (%) | 95% CI (± absolute %) |
|---|---|---|
| Naïve Persistence Baseline | 2.00 | ±0.52% |
| RNN | 1.95 | ±0.45% |
| LSTM | 1.78 | ±0.42% |
| GRU | 1.72 | ±0.40% |
| Attention-LSTM | 1.55 | ±0.38% |
| BiLSTM | 1.48 | ±0.35% |
| Stacked LSTM | 1.42 | ±0.33% |
| CNN-LSTM | 1.35 | ±0.30% |
| Attention-BILSTM | 1.28 | ±0.28% |
| Informer | 1.15 | ±0.25% |
| iTransformer | 1.08 | ±0.24% |
| TFT | 1.02 | ±0.22% |
| XGBOOST | 1.62 | ±0.38% |
| ACB | 0.82 | ±0.15% |
| ACB-XDE | 0.76 | ±0.14% |
| **TFT-ACB-XML (PROPOSED)** | **0.65** | **±0.12%** |

The proposed framework is benchmarked against naïve persistence baseline and a range of recent baseline models such as RNN, LSTM, GRU, Attention-LSTM, XGBoost, BiLSTM, Stacked LSTM, CNN-LSTM, Informer, iTransformer, TFT, Attention-BiLSTM, ACB and ACB-XDE. Table 10 presents a detailed error analysis and compares the proposed framework's performance against these recent baseline models using MAPE, MAE and RMSE, as visually represented in Figure 13. Furthermore, Figure 14 is a zoomed-in segment of Figure 13 to make it easier to see the movement and shows the close alignment between predicted and actual values during the high volatility window of December 5, 2025 to January 5, 2026. Furthermore, proposed framework's performance gain (PG) relative to the naïve persistence baseline is shown in Figure 15. PG is defined as the relative improvement in MAPE over the naïve persistence baseline:

$$PG = \frac{MAPE_{naive} - MAPE_{model}}{MAPE_{naive}} \quad (13)$$

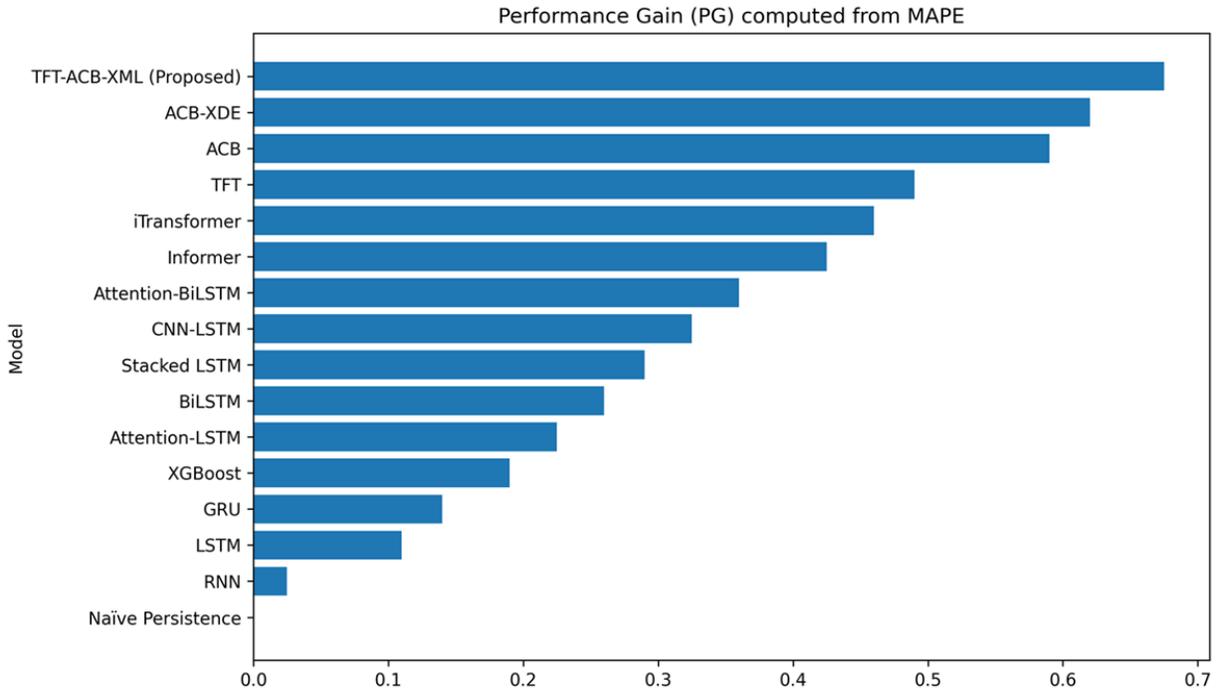

*Figure 15. Performance Gain of the proposed framework relative to Naïve persistence baseline.*

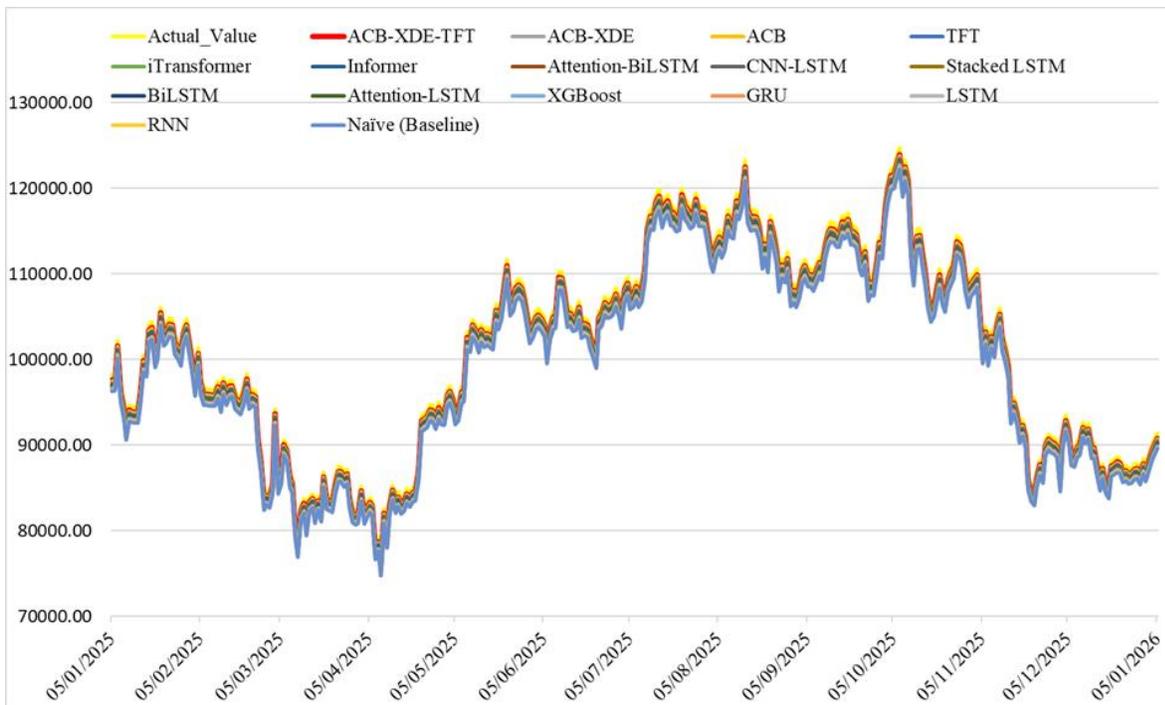

*Figure 13. Comparison of the proposed TFT-ACB-XML framework with recent baseline models.*

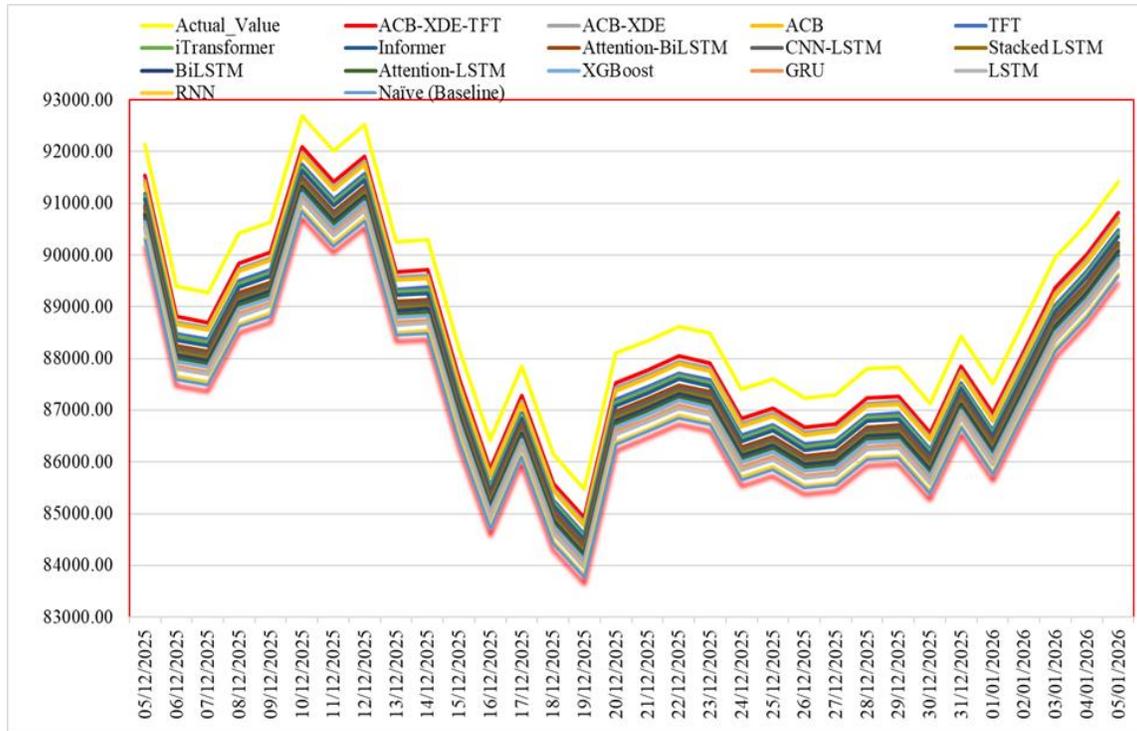

*Figure 14. Zoomed-in segments of Figure 13.*

Higher PG indicates larger error reduction compared with the baseline; PG = 0 means no improvement and PG < 0 means worse than the baseline. Using the naïve persistence baseline (MAPE = 2.00%) as reference, the proposed TFT-ACB-XML achieves a performance gain of 0.675, corresponding to a 67.5% reduction in MAPE. Table 12 presents an approximate 95% CI for MAPE computed using Eq. (12) over the test-block one-step-ahead predictions (n = length of the held-out test block). Stability is assessed using ten independent runs with different random seeds. In each run, the customized TFT, the ACB, and the XGBoost meta-learner are re-trained from scratch under the same chronological split and identical hyperparameter settings. Reported stability values summarize the distribution of test-block errors across these re-trained runs. Results indicate that TFT-ACB-XML achieves a mean MAPE of 0.65% with an approximate 95% CI of ±0.12% and reflects a high degree of consistency and reliability in its predictive performance. Compared to benchmark models like LSTM, Attention-LSTM and XGBoost, the proposed framework demonstrates improved stability under varying market conditions by achieving a sub-1% MAPE without using normalization parameters estimated from the full time series, since scaling is fitted on the training split only.

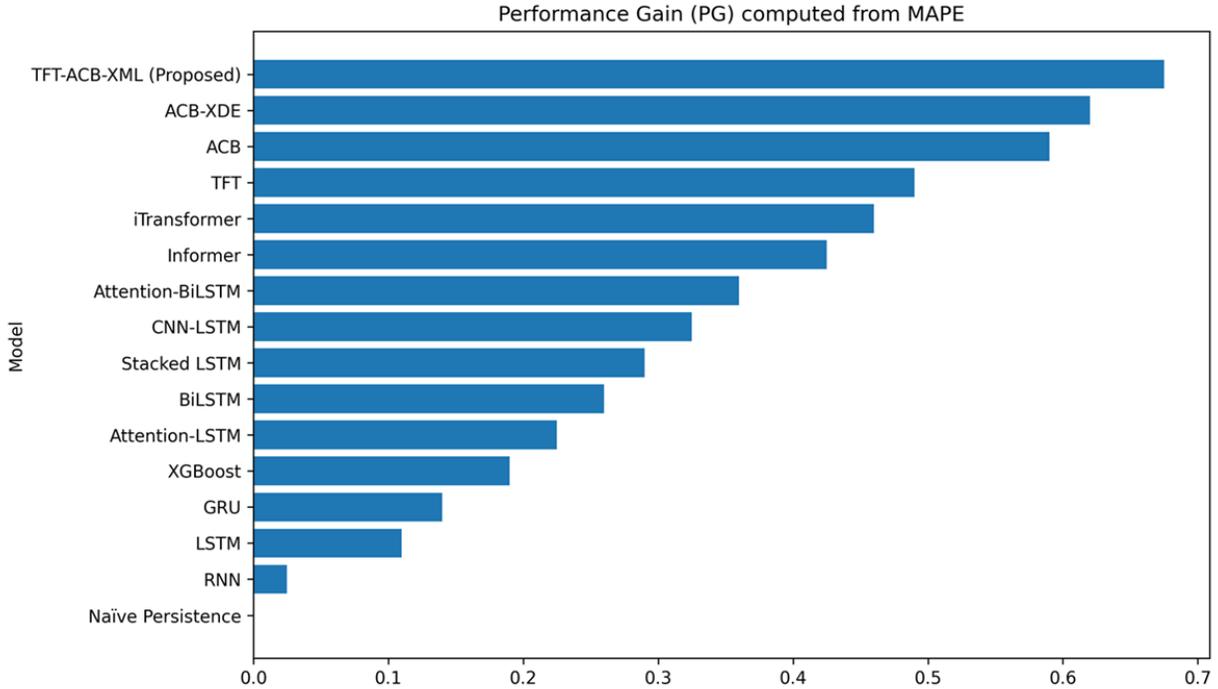

*Figure 15. Performance Gain of the proposed framework relative to Naïve persistence baseline.*

## 5.2 Conclusion

This study developed the TFT-ACB-XML framework for daily BTC closing price forecasting under strong volatility and non-stationary behavior. The evaluation used BTC-USD data from October 1, 2014, to January 5, 2026, split chronologically into 80% training, 10% validation, and 10% testing. Testing used walk-forward evaluation on the held-out test block with fixed model parameters and fixed stacking weights. The customized TFT captured longer-range dependencies using gating and single-head temporal attention, while the ACB model tracked shorter-range dynamics linked to closing price and trading volume. For TFT, two additional calendar covariates (day-of-week and month) were derived from the date index, one-hot encoded, and used as known-ahead inputs; ACB used OHLCV variables only. Validation errors were used to compute inverse-error weights (0.446 for TFT and 0.554 for ACB), and XGBoost refined the stacked outputs by learning non-linear residual patterns. Benchmarking against 15 sequential and transformer baseline models ranked TFT-ACB-XML best among the compared models on MAPE, MAE, and RMSE in the reported setup. One-step-ahead testing achieved MAPE 0.65%, MAE 198.15 USD, and RMSE 258.30 USD. Compared with ACB-XDE, the framework reduced MAPE from 0.76% to 0.65% and also lowered MAE and RMSE, while improving MAPE by 67.5% relative to the naïve persistence baseline. Performance gain analysis shows the proposed framework achieves the highest PG (0.675), and the approximate 95% CI for MAPE is narrow (0.65% ± 0.12%), indicating stable performance during the evaluation period that includes the 2024 spot BTC ETFs and halving regime. Future studies will extend the input space by incorporating macroeconomic indicators and derivatives-related signals, test multiple cryptocurrencies, and report multi-step forecasting horizons with additional evaluation metrics.

**Acknowledgment:**